\documentclass{article}
\usepackage{spconf,amsmath,graphicx,hyperref, bm}
\usepackage{amssymb} 

\usepackage{comment}
\usepackage{multirow}
\usepackage{stfloats}
\usepackage{booktabs}
\usepackage{graphicx} 
\usepackage{makecell} 

\title{Frame Sampling Strategies Matter: A Benchmark
for small vision language models}
\name{
  Marija Brkic\textsuperscript{1},
  Anas Filali Razzouki\textsuperscript{1, 2},
  Yannis Tevissen\textsuperscript{2},
  Khalil Guetari\textsuperscript{2},
  Mounim A. El-Yacoubi\textsuperscript{1}
}

\address{
  \textsuperscript{1}Télécom SudParis, Institut Polytechnique de Paris, France \\
  \textsuperscript{2}Moments Lab Research, France
}

\begin{document}
%
\maketitle
\begin{abstract}
Comparing vision language models (VLMs) on videos is particularly complex, as the performances is jointly determined by the model’s visual representation capacity and the frame-sampling strategy used to construct the input. Current video benchmarks are suspected to suffer from substantial frame-sampling bias, as models are evaluated with different frame selection strategies. In this work, we propose the first controlled frame-sampling  benchmark of state-of-the-art small VLMs (SVLMs) for video question-answering. Our results confirm the suspected bias and highlight both data-specific and task-specific behaviors of SVLMs under different frame-sampling techniques.
By open-sourcing our benchmarking code, we provide the community with a reproducible and unbiased protocol for evaluating video VLMs and emphasize the need for standardized frame-sampling strategies tailored to each benchmarking dataset in future research.

\end{abstract}
\begin{keywords}
vision-language models, frame-sampling bias, video question-answering, benchmark, controlled frame-sampling benchmark
\end{keywords}
\section{Introduction}
\label{sec:intro}
Multimodal learning has advanced rapidly in recent years, with models now able to jointly process and reason over images, text, and audio. Within this field, Vision-Language Models (VLMs) integrate visual and textual prompt inputs to generate text grounded in the visual input, enabling tasks such as visual question answering (VQA), object localization, and document understanding \cite{li2025surveystateartlarge}. 

Large VLMs deliver strong, state-of-the-art results in multimodal reasoning and tasks like VQA and document understanding. At the same time, advances in efficiency, scalability, and the ability to operate on devices with limited resources have motivated a shift away from large-parameter VLMs toward more compact designs, called Small Vision-Language Models (SVLMs, up to 4B params) \cite{Shinde2025Survey, patnaik2025svlm_survey}. They often employ knowledge distillation, compact architectures, and curated datasets optimized for efficiency \cite{marafioti2025smolvlm}.  

VLMs typically rely on three components: a pretrained vision encoder, a pretrained language model (LLM), and a connector that maps visual features to the language embedding space, commonly instantiated as a multi-layer perceptron (MLP). SVLMs use lighter-weight designs for each component.

Among the prominent SVLM families are Qwen2.5-VL-3B, Ovis2-2B, InternVL3-2B, and SmolVLM2-2.2B \cite{Qwen2.5-VL, lu2024ovis,InternVL3_2025,HuggingFace2025SmolVLM2}. All follow the vision–connector–LLM pattern with ViT backbones, yet each pursues a distinct lightweight strategy. Qwen2.5-VL-3B uses a ~675M DFN-based ViT with 2D-RoPE and a token-compression MLP before the Qwen2.5-3B decoder. Ovis2-2B retains a CLIP-style ViT \cite{Radford2021CLIP} but replaces plain projection with probabilistic visual tokenization over a learnable visual embedding table, yielding structured visual embeddings aligned to the text embedding scheme, paired with a small Qwen2-1.7B LLM \cite{qwen2}. InternVL3-2B adopts an InternViT-300M encoder with pixel unshuffle and variable visual position encoding, trained natively with Qwen2.5-1.5B to align modalities autoregressively rather than through contrastive pretraining. SmolVLM2-2.2B prioritizes efficiency by coupling a SigLIP-SO400M \cite{zhai2023sigmoidlosslanguageimage} vision encoder with pixel-shuffle token reduction and an MLP projection into SmolLM2-1.7B \cite{allal2025smollm2smolgoesbig}, enabling long contexts and edge-friendly compute.

These SVLMs report competitive, state-of-the-art results on video-QA benchmarks. However, cross-model comparisons are confounded by divergent frame-sampling protocols. Although most adopt uniform sampling at a fixed frame rate (FPS), the effective number of frames per video varies substantially across studies, which obscures the true performance of each SVLM. For instance, Qwen2.5 samples at 2 FPS with a cap of 768 frames, SmolVLM uses 1 FPS with up to 50 frames, and Ovis2 and InternVL3 apply uniform sampling with even fewer frames. Consequently, the reported results are not directly comparable. 

Although uniform sampling is widely used, it remains a simple approach to sample frames and can miss key dynamics. Adaptive frame sampling techniques have been proposed, including prediction-based selection \cite{yu2024framevoyager}, shot-boundary detection \cite{baraldi2017hierarchical}, similarity scoring between queries and frames \cite{liu2025boltboostlargevisionlanguage} and frame embeddings based extraction \cite{li2025maxinfo}. To the best of our knowledge, there are no existing benchmarks that compare SVLMs under the same frame sampling strategies.

To address these limitations, we propose a new standardized benchmark by comparing state-of-the-art SVLMs, without any frame sampling bias under two protocols: Standard sampling and adaptive sampling. In the standard sampling, we used uniform-FPS sampling strategy and single-frame sampling. In the adaptive setting, we leverage MAXInfo algorithm and also propose a CSTA-based method that leverages the pretrained CSTA model \cite{son2024csta}, originally developed for video summarization. We open-source the code and selected frames under different settings so that the community can benchmark other video understanding approaches: \href{https://github.com/momentslab/frame-sampling}{https://github.com/momentslab/frame-sampling}.

\section{Method}
\label{sec:method}

\subsection{Frame sampling strategies}
\label{ssec:sampling}

\subsubsection{Standard sampling}
\label{sssec:standard}
We categorize standard sampling into two methods: uniform-FPS sampling and single-frame sampling.  
In uniform-FPS sampling, frames are extracted at a fixed rate of \(r\) frames per second (e.g., for \(r=2\), two frames are taken each second). An upper cap \(N_{\max}\) is applied to satisfy the compute budget, and a lower bound \(N_{\min}\) ensures coverage for very short videos.  
In single-frame sampling, exactly one frame is selected, either the center frame or the first frame of the video. Although this approach cannot capture temporal dynamics, it provides insight into model performance in the absence of motion cues.

\subsubsection{MaxInfo}
\label{sssec:maxinfo}
MaxInfo algorithm was proposed in \cite{li2025maxinfo} as a training-free method to dynamically sample the most informative frames using the Rectangular Maximum Volume algorithm (MaxVol) \cite{MikhalevOseledets2015}. MaxInfo works as follows: 

First, $n$ initial frames are sampled uniformly from a video. For each of these frames, visual token embeddings are calculated using CLIP-ViT-B/32. [CLS] embeddings are retained and stacked to obtain a matrix \(\mathbf{Q} \in \mathbb{R}^{n \times d}\), where \(d\) denotes the embedding dimension.

Then a truncated SVD is performed to reduce the dimension of $\mathbf{Q}$, resulting in $\mathbf{Q}_s \in \mathbb{R}^{n \times s}$. Next, the MaxVol algorithm \cite{MikhalevOseledets2015} is applied to extract the most informative subspace from $\mathbf{Q}_s$, resulting in a subset that contains $p$ frames. We also set an upper cap $N_{\max}$. If $p > N_{\max}$, then $N_{\max}$ frames are uniformly selected from the $p$ frames.

\subsubsection{CSTA}
\label{sssec:csta}

Video summarization aims to condense a video while preserving its essential content, commonly formulated as keyshot selection under a length budget. Son et \textit{al.} \cite{son2024csta} proposes a CNN-based SpatioTemporal Attention (CSTA) architecture to simultaneously capture spatial and temporal relationships using a 2D-CNN. 

We adapt the pretrained CSTA model for video understanding. First, $n$ initial frames are uniformly sampled from the video. Each frame is then embedded with \emph{GoogLeNet} \cite{Szegedy2015GoingDeeper}, and the embeddings are fed to an attention module that produces an attention map over the concatenated frames. A mixing module combines these maps with the frame features and passes the result to a classifier, which assigns an importance score to each frame.

We retain the top 15\% of frames (150 frames) with the highest scores. To control the budget, we also impose an upper cap $N_{\max}$: if $N_{\max} < 150$, we uniformly subsample $N_{\max}$ frames from this set.

\subsection{Proposed benchmark}
\label{ssec:benchmark}

VideoMME \cite{fu2025video} and MVBench \cite{li2024mvbench} are the two benchmark datasets used to evaluate frame sampling strategies. Both are large-scale, multi-task video understanding benchmarks that use multiple-choice QA to assess capabilities such as temporal reasoning, spatial relations, and action recognition. VideoMME contains open-domain videos ranging from 11 seconds to 1 hour, with an average duration of 17 minutes per video.  In contrast, MVBench is composed of short clips, averaging roughly 16 seconds and typically less than one minute.

\subsection{Inference setup}
\label{ssec:setup}
We evaluate four small vision--language models, namely Qwen2.5-VL-3B, Ovis2-2B, InternVL3-2B, and SmolVLM2-2.2B, on two video understanding benchmarks: VideoMME and MVBench. For each model, inference is conducted on video frames obtained using four sampling strategies:  single-frame sampling, uniform FPS sampling, and two adaptive methods : MaxInfo and CSTA.

All models are evaluated using the official prompts provided by each benchmark, ensuring consistent prompt control across experiments. We rely exclusively on fixed pretrained checkpoints and deterministic inference. As a result, there is no variability induced by random seeds, and the dominant source of uncertainty arises from the finite size of the evaluation datasets. Accordingly, we estimate standard errors using a binomial approximation that depends solely on the dataset size.

For uniform FPS sampling, we use a frame rate of $r = 2$, corresponding to two frames per second, with a minimum of $N_{\min} = 4$ and a maximum of $N_{\max} = 96$ frames per video. This configuration is denoted as FPS:2:4:96. In the single-frame setting, we evaluate two variants by selecting either the center frame or the first frame of each video.

For the adaptive sampling strategies MaxInfo and CSTA, we first construct an initial pool of $n = 1000$ uniformly sampled frames to balance temporal coverage and computational cost. This choice is particularly important for MaxInfo, as it involves SVD and MaxVol operations with computational complexity that scales approximately quadratically with the number of frames. Each method then selects an informative subset of frames from this pool, with the final selection capped at $N_{\max} = 96$ frames, matching the upper bound used in the uniform FPS setting.

The choice of $N_{\max} = 96$ is dictated by the visual token capacity of the smallest model considered in our study, namely SmolVLM. Fixing this upper limit across all experiments enforces a uniform token budget for all models, thereby enabling fair and controlled cross-model comparisons. All experiments were conducted under a modest computational budget, corresponding to approximately two weeks of runtime on four NVIDIA A10G GPUs.
\section{Results}
\label{sec:results}

\subsection{Performance Analysis}

Table \ref{tab:results} summarizes SVLMs performances under several frame sampling strategies on the VideoMME and MVBench benchmarks. On VideoMME, uniform-FPS sampling is best for every model. Qwen2.5 achieved the highest accuracy at 60.9\%, and all SVLMs benefit from evenly spaced frames. By contrast, on MVBench, the optimal strategy is model-specific. Qwen2.5 achieves the best performance with MaxInfo (68.4\%). SmolVLM2 and Ovis2 perform best with uniform-FPS (51.0\% and 65.8\%, respectively). InternVL3 reaches its highest score with CSTA (65.8\%).

These patterns reflect both dataset design and model characteristics. VideoMME rewards consistent temporal sampling that captures fine-grained changes. MVBench spans more diverse tasks, some solvable from a few salient frames and others that benefit from adaptive selection or full temporal coverage.

\begin{figure}[h]
    \centering
    \includegraphics[width=1\linewidth]{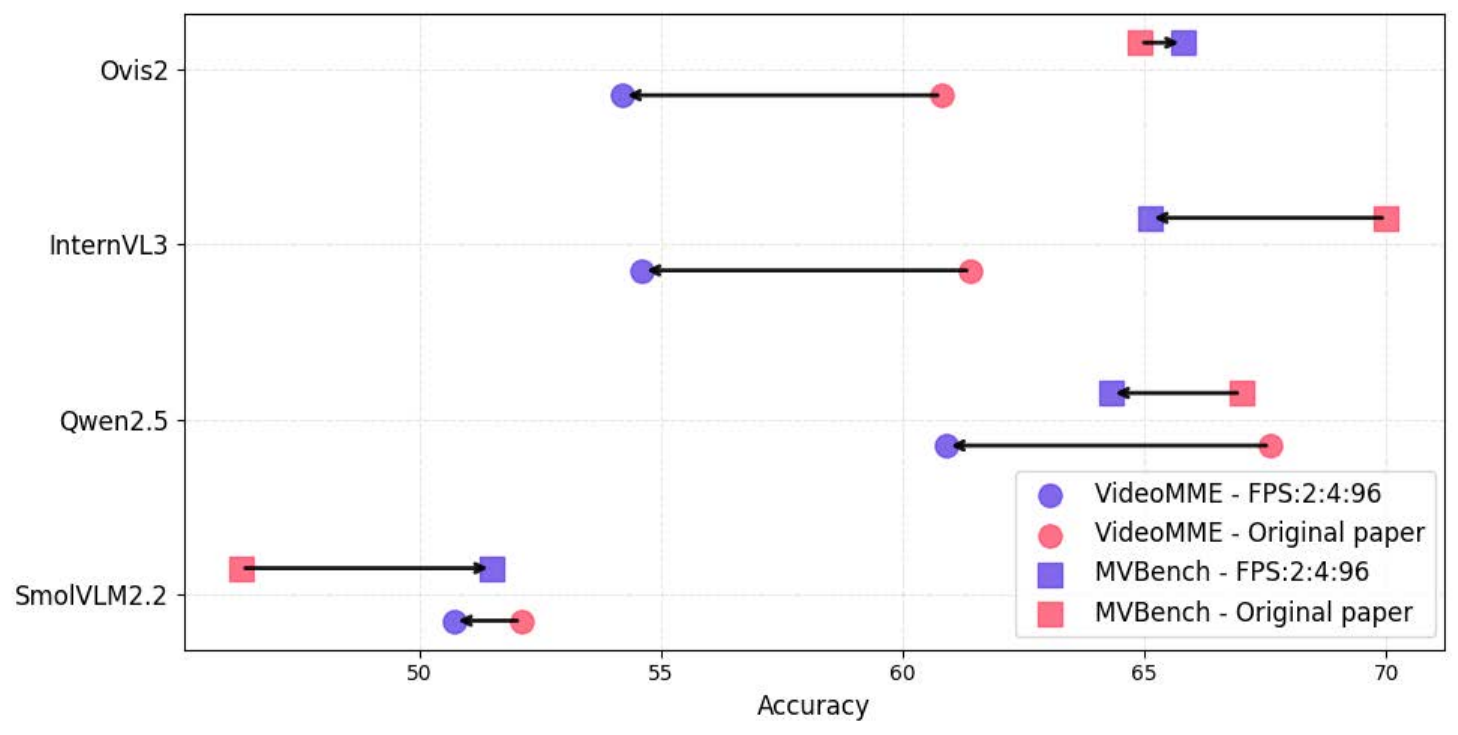}
    \caption{Comparison between results reported in the original papers and our re-evaluation under a unified sampling setup for FPS uniform sampling.}
    \label{fig:reported}
\end{figure}

Figure~\ref{fig:reported} compares our uniform-FPS results with literature reported scores for these SVLMs on VideoMME and MVBench. We can observe that once sampling strategy is standardized, performance can shift in either direction, indicating strong sensitivity to frame selection. The largest shift is observed for InternVL3 on MVBench, where accuracy dropped from about 70\% to 65\%. This drop is likely due to a distribution shift: InternVL3 is evaluated with a fixed number of uniformly spaced frames (16 frames), whereas our fixed 2 fps regime yields 4 to 96 frames, under-samples fast actions in short clips. In contrast, Qwen2.5 shows little difference, as our uniform-FPS setting is close to that reported in the literature (they used FPS:2:4:768, while we used FPS:2:4:96). Using a common token budget and a single sampling policy removes this confound and makes cross-model differences more meaningful.

\begin{table}[h]
  \centering
  \scriptsize                     
  \setlength{\tabcolsep}{2pt}     
  \renewcommand{\arraystretch}{1} 
  \caption{Comparison of VideoMME and MVBench results across frame sampling strategies. FPS (fps:2:4:96): uniform sampling at 2 fps. Max Info (maxinfo:1000:96) and CSTA (csta:1000:96): adaptive selection of up to 96 frames from 1000 inputs. Standard errors are reported using binomial approximation to estimate the uncertainty inherent to the finite evaluation sets.}
  \label{tab:results}

  \resizebox{\columnwidth}{!}{%
  \begin{tabular}{@{}l|ccccc|ccccc@{}}
    \toprule
    \multirow{2}{*}{Model} &
      \multicolumn{5}{c|}{VideoMME ($\pm$1.66\%)} &
      \multicolumn{5}{c}{MVBench ($\pm$0.76\%)} \\
    \cmidrule(lr){2-6} \cmidrule(lr){7-11}
     & First & Center & FPS & Max Info & CSTA
     & First & Center & FPS & Max Info & CSTA \\
    \midrule
    Qwen2.5    & 41.9 & 43.5 & \textbf{60.9} & 57.5 & 57.9 
                & 47.5 & 50.9 & 64.3 & \textbf{68.4} & 64.0 \\
    SmolVLM2    & 36.2 & 38.4 & \textbf{50.7} & 48.7 & 49.2 
                & 42.9 & 45.0 & \textbf{51.5} & 51.0 & 50.5 \\
    InternVL3 & 37.9 & 42.8 & \textbf{54.6} & 53.6 & 53.3 
                 & 44.9 & 48.4 & 65.1 & 63.1 & \textbf{65.8} \\
    Ovis2 & 40.8 & 42.1 & \textbf{54.2} & 52.5 & 53.3 
              & 47.3 & 51.8 & \textbf{65.8} & 62.8 & 62.3 \\
    \bottomrule
  \end{tabular}%
  }
\end{table}

\begin{figure}[h]
    \centering
    \includegraphics[width=\linewidth]{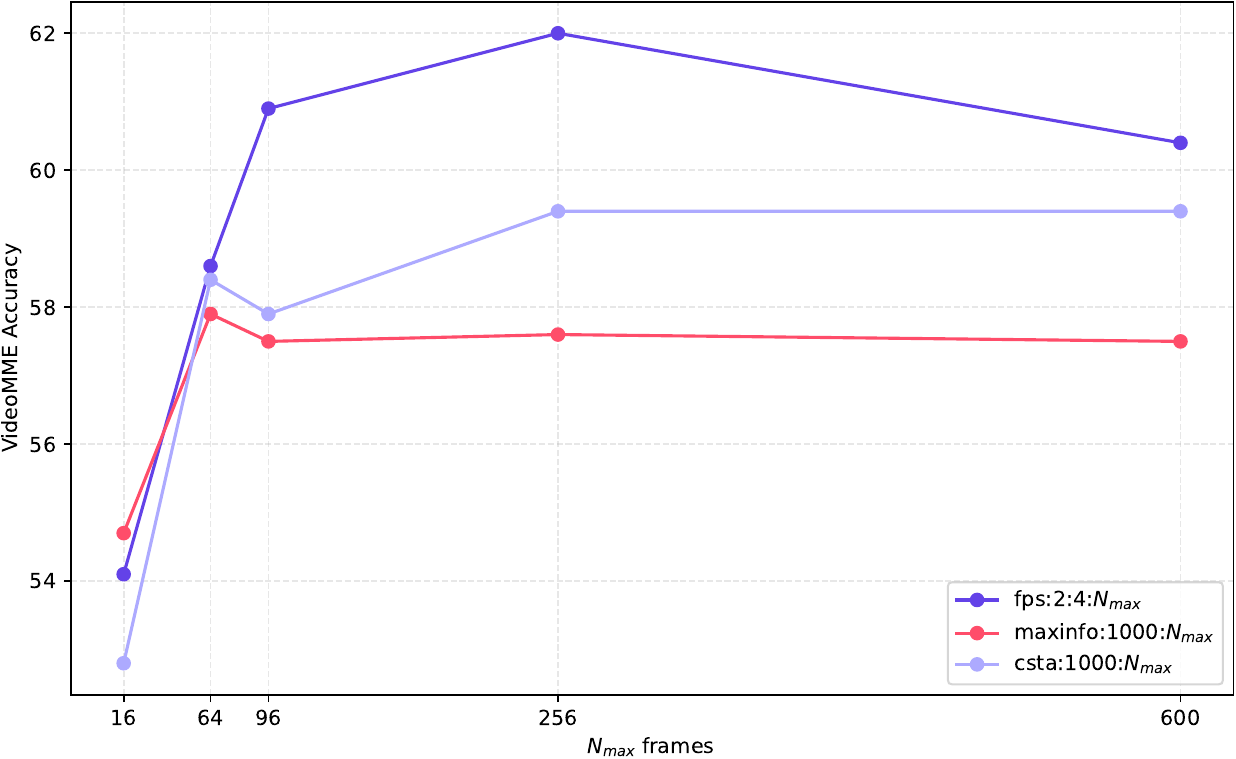}
    \caption{Evolution of Qwen2.5 accuracy on VideoMME when changing the maximum number of frames used after each sampling strategy}
    \label{fig:frame_ablation}
\end{figure}

We also performed an ablation study with Qwen2.5 on VideoMME to evaluate the impact of changing the maximum number of frames within the different sampling techniques (FPS, CSTA and MaxInfo). Configurations with 16, 64, 96, 256 and 600 frames were tested and reported in Figure \ref{fig:frame_ablation}.

The results show that uniform-FPS sampling benefits the most from larger frame counts. 
The accuracy rises sharply from 16 to 256 frames, peaking around 62\% at 256, before slightly dropping at 600 frames. 
This indicates that FPS gains from additional temporal coverage, but excessive frames may introduce redundancy or noise that weakens performance.

Accuracy with MaxInfo rises up to 64 frames and then remains nearly constant from 96 to 600, while CSTA increases steadily from 16 to 256 frames before plateauing.

\begin{figure*}[h]
    \centering
    \includegraphics[width=\linewidth]{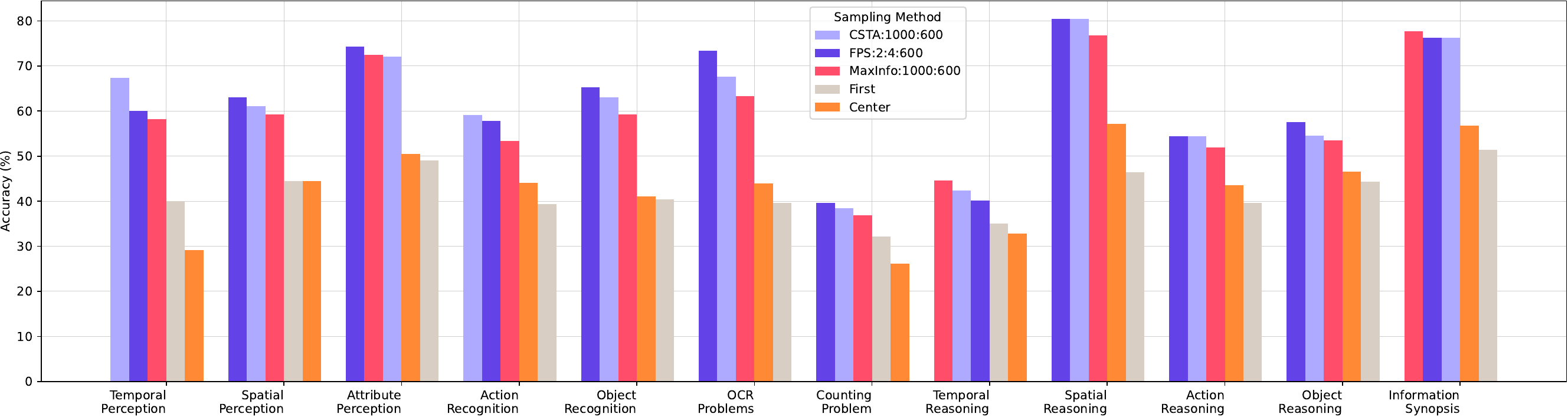}
    \caption{Distribution of Qwen2.5 performances over all VideoMME tasks with various frame sampling strategies.}
    \label{fig:qwen2_tasks}
\end{figure*}

Additionally, in Figure~\ref{fig:qwen2_tasks}, we present Qwen2.5 results with $N_{\max}=600$ across the VideoMME task groups. Overall, uniform FPS outperforms the other strategies, with few exceptions: CSTA leads in temporal perception and action recognition, while MaxInfo is slightly better in temporal reasoning and information synopsis. 

\subsection{Runtime Analysis}
Table \ref{tab:inference_time} presents the wall-clock inference time (in hours) for VideoMME and MVBench across different frame-sampling strategies, corresponding to the experiments whose results are reported in Table \ref{tab:results}.  Single-frame inference is computationally negligible for all models, remaining below 0.36 h in every configuration. This highlights the low overhead of per-video inference when minimal temporal context is used.

In contrast, uniform FPS sampling incurs a substantial increase in runtime, with inference times ranging from approximately 5–7 h on VideoMME and 1.6–2.4 h on MVBench. This increase reflects the processing of up to 96 frames per video, confirming the near-linear relationship between frame count and inference cost.

The most computationally demanding strategies are Max Info and CSTA. These methods require an expensive pre-selection stage, leading to inference times exceeding 21 h on VideoMME and 6–12 h on MVBench across all models. Importantly, Max Info and CSTA exhibit comparable runtimes, indicating that their computational burden is dominated by the initial dense frame processing.

\begin{figure}[h]
    \centering
    \includegraphics[width=\linewidth]{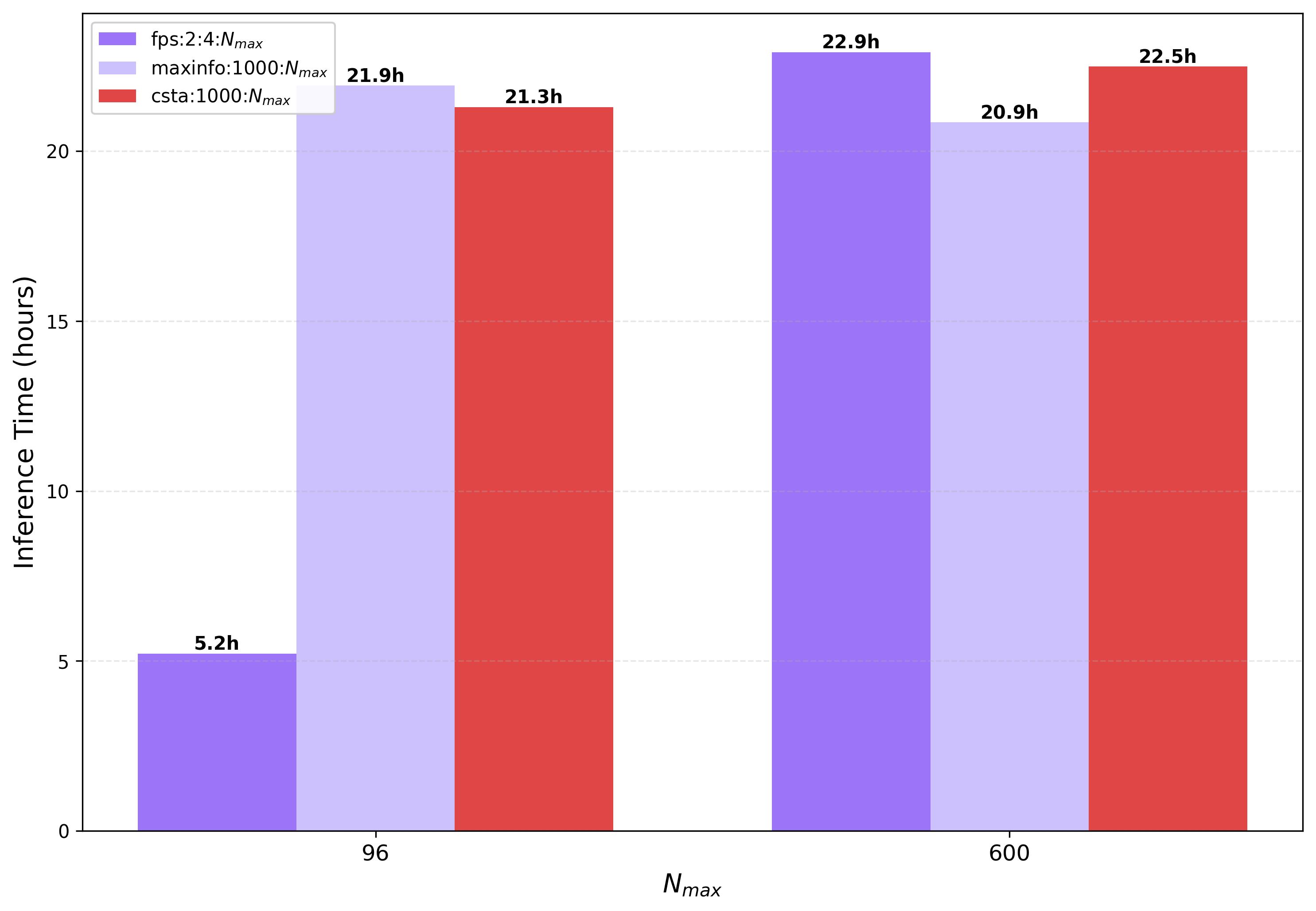}
    \caption{Impact of maximum frame limit on inference time}
    \label{fig:inference_qwen}
\end{figure}

We also report in Figure~\ref{fig:inference_qwen} the inference runtime, focusing on Qwen2.5 performance on VideoMME as $N_{\max}$ increases from 64 to 600.
 The results show that dynamic sampling methods can become particularly useful when the frame budget is large (e.g., $N_{\max}=600$). In this regime, the additional preprocessing cost incurred during frame preselection allows these methods to reduce the number of frames ultimately processed by the SVLM. In contrast, FPS directly processes up to 600 frames without any preselection, leading to similarly high inference times at large $N_{\max}$.

\begin{table}[h]
  \centering
  \scriptsize                     
  \setlength{\tabcolsep}{2pt}     
  \renewcommand{\arraystretch}{1} 
  \caption{Inference time comparison (in hours) across frame sampling strategies. FPS (fps:2:4:96): uniform sampling at 2 fps. Max Info (maxinfo:1000:96) and CSTA (csta:1000:96): adaptive selection of up to 96 frames from 1000 inputs.}
  \label{tab:inference_time}

  \resizebox{\columnwidth}{!}{%
  \begin{tabular}{@{}l|cccc|cccc@{}}
    \toprule
    \multirow{2}{*}{Model} &
      \multicolumn{4}{c|}{VideoMME} &
      \multicolumn{4}{c}{MVBench} \\
    \cmidrule(lr){2-5} \cmidrule(lr){6-9}
     & Single Frame & FPS & Max Info & CSTA
     & Single Frame & FPS & Max Info & CSTA \\
    \midrule
    Qwen2.5    & 0.23 & 5.22 & 21.93 & 21.30 
                & 0.31 & 2.05 & 11.04 & 7.03 \\
    SmolVLM    & 0.33 & 5.41 & 21.30 & 22.06 
                & 0.25 & 1.67 & 9.97 & 6.49 \\
    InternVL   & 0.26 & 6.27 & 23.10 & 22.56 
                 & 0.36 & 1.98 & 10.85 & 7.04 \\
    Ovis       & 0.21 & 6.90 & 23.49 & 23.10 
              & 0.34 & 2.38 & 11.91 & 7.67 \\
    \bottomrule
  \end{tabular}%
  }
\end{table}

\section{Discussion}
\label{sec:disc}

We present a standardized benchmark of state-of-the-art SVLMs that controls frame-sampling bias under two protocols: standard sampling and adaptive sampling. On VideoMME, Qwen2.5 attains the highest accuracy with uniform-FPS, outperforming other SVLMs by at least 6\%. On MVBench, Qwen2.5 also leads with MaxInfo, exceeding other SVLMs by at least 3\%. Ovis2 and InternVL3 reach comparable accuracies, both outperforming SmolVLM2 across the two benchmarks.

These results highlight the importance of the model’s training data distribution and its alignment with each benchmark. MVBench emphasizes short, motion-rich clips and prompts that stress temporal reasoning, whereas Video-MME spans broader durations and a wider mix of skills. As a result, models whose pretraining favors brief, dynamic sequences may align better with MVBench, while those tuned for longer contexts may fare better on VideoMME. Model size may also contribute: although all models are small (up to 4B parameters), Qwen2.5 (3.7B) is larger than Ovis2 and InternVL3 (2B each).

For sampling strategies, uniform FPS is the strongest approach on VideoMME across all evaluated models. In contrast, performance on MVBench is model dependent. This pattern indicates that VideoMME rewards steady temporal coverage: evenly spaced frames preserve event order and reduce the risk of missing brief, repeated motions. On MVBench, performance is model dependent: some models, including Qwen2.5 and InternVL3, benefit from adaptive sampling. However, the gain comes with higher compute and time costs, and adaptive sampling is roughly three to five times slower than standard sampling because of preprocessing for frame selection. Single-frame sampling (center or first) underperforms all other strategies because it provides too little motion information and context for reliable responses.

Another important point is the comparison between literature-reported results and our controlled evaluation. Overall, controlling for frame-sampling bias is effective: it reveals discrepancies relative to reported scores. On MVBench, prior reports place Ovis2 below InternVL3 and Qwen2.5 (64.9\% vs. 70.0\% and 67.0\%), partly because Qwen2.5 used uniform-FPS sampling (up to 768 frames) and InternVL3 evaluated 16 uniformly spaced frames, whereas Ovis2 was evaluated with only a few frames (up to 12). Under our controlled protocol, the ranking changes: when all models receive the same input frames, Ovis2 performs comparably (Ovis2 65.8\% vs. InternVL3 65.0\% and Qwen2.5 64.3\%).

On VideoMME, Qwen2.5 performance scales with the frame budget: increasing $N_{\max}$ from 16 to 256 yields marked gains across sampling strategies, 
reflecting improved temporal coverage. Beyond 256 frames, accuracy plateaus or decreases slightly, likely due to redundancy and noise. We also find that dynamic sampling could becomes particularly useful when the frame budget is large, since the cost incurred during frame preselection allows these methods to reduce the number of frames ultimately processed by the SVLM, making the final inference time comparable to that of uniform sampling. 

Furthermore, we found that within VideoMME, uniform FPS sampling outperformed the other strategies on most tasks, except for temporal tasks where adaptive strategies achieved better performance. This can be attributed to the fact that temporal tasks rely on capturing fine-grained motion dynamics and event ordering. Adaptive strategies are better suited for this, as they prioritize frames with higher temporal variability, thereby preserving critical motion cues.

We acknowledge that this study is restricted to small VLMs due to hardware constraints. Also, this evaluation focuses on video question-answering, thus capturing only part of the broader video understanding landscape. Finally, only a subset of adaptive frame sampling strategies is considered, and the results should be interpreted within this methodological frame.

\section{Conclusion}
\label{sec:conclusion}
In this work, we present the first controlled benchmark of four small VLMs on video question answering under multiple frame-sampling strategies. We found substantial gaps between literature-reported scores, which use inconsistent frame sampling, and our controlled evaluations. These discrepancies obscure cross-model differences, whereas a standardized sampling protocol enables fair and reproducible comparisons. 

For future work, one could search for more refined adaptive strategies that could deliver consistent gains across tasks and benchmark data, building on our finding on temporal tasks. A more effective strategy is to integrate frame selection into VLM training, enabling the model to jointly optimize which frames to sample for the target task. Such methods would improve performance and help ensure fair cross-model comparisons.

\bibliographystyle{IEEEbib}
\bibliography{main}

\end{document}